\documentclass[10pt, a4paper]{article}

\usepackage{booktabs}
\usepackage{tikz}
\usetikzlibrary{positioning} 
\usepackage{minted}
\usepackage{xcolor}
\usepackage{enumerate}

\usepackage[final]{lrec2026}

\title{Do LLMs Judge Distantly Supervised Named Entity Labels Well? Constructing the \texttt{JudgeWEL} Dataset}

\name{Alistair Plum\textsuperscript{1}, Laura Bernardy\textsuperscript{1}, Tharindu Ranasinghe\textsuperscript{2}} 

\address{\textsuperscript{1}University of Luxembourg, Luxembourg \\
         \textsuperscript{2}Lancaster University, UK \\
         \texttt{\{alistair.plum, laura.bernardy\}@uni.lu}\\
         \texttt{t.ranasinghe@lancaster.ac.uk} \\}

\abstract{
We present judgeWEL, a dataset for named entity recognition (NER) in Luxembourgish, automatically labelled and subsequently verified using large language models (LLM) in a novel pipeline. Building datasets for under-represented languages remains one of the major bottlenecks in natural language processing, where the scarcity of resources and linguistic particularities make large-scale annotation costly and potentially inconsistent. To address these challenges, we propose and evaluate a novel approach that leverages Wikipedia and Wikidata as structured sources of weak supervision. By exploiting internal links within Wikipedia articles, we infer entity types based on their corresponding Wikidata entries, thereby generating initial annotations with minimal human intervention. Because such links are not uniformly reliable, we mitigate noise by employing and comparing several LLMs to identify and retain only high-quality labelled sentences. The resulting corpus is approximately five times larger than the currently available Luxembourgish NER dataset and offers broader and more balanced coverage across entity categories, providing a substantial new resource for multilingual and low-resource NER research.
\\ \newline  }

\begin{document}

\maketitleabstract

\section{Introduction}\label{sec:intro}
Recent advances in large language models (LLMs) have redefined the scope of natural language processing (NLP), shifting attention from classification and sequence-labelling tasks to increasingly complex generative applications. This transformation has enabled conversational systems that integrate reasoning, translation, and summarisation within a single multi-task architecture \cite{brown2020language,touvron2023llama,jiang2023mistral}. Yet, the reach of these technologies remains uneven. For many less-resourced and under-represented languages, the benefits are limited: they continue to lack fundamental resources such as labelled corpora that support even basic supervised learning tasks, including sequence labelling and text classification \cite{joshi2020state,adelani2022masakhaner}.

Building datasets for such languages remains particularly challenging. Obstacles typically include a scarcity of written material, limited linguistic expertise, and the absence of sustained funding, all of which reinforce a cycle of under-representation. When coupled with limited prior research, these constraints make it difficult to bootstrap the resources necessary for progress \cite{hettiarachchi-etal-2025-overview}. In these circumstances, creative and data-efficient strategies for dataset construction are essential. This is the case for Luxembourgish, the national language of Luxembourg and one of three official languages alongside German and French. Despite its official status, Luxembourgish is the only national language of an EU member state not represented at the EU administrative level, thus remaining absent from standardised resources such as Europarl \cite{koehn2005europarl}.

Luxembourgish is under-represented in NLP \cite{joshi2020state} compared to its larger neighbours German and French, though it shares deep historical and structural ties with both. It evolved from the Moselle Franconian dialect \cite{Gilles2019}, displaying strong affinities with German while integrating extensive French lexical and grammatical. Code-switching and borrowing are common, particularly in informal written communication, giving the language a dynamic character that further complicates computational treatment. Despite these challenges, the language’s roughly 400,000 speakers \cite{Gilles2019}, ongoing standardisation efforts, and renewed digital interest make this an opportune moment to improve NLP for Luxembourgish. 

In this paper, we present a new methodology for the automatic creation of a named entity recognition (NER) dataset. The methodology is designed to achieve broad and reliable coverage with minimal human supervision. To overcome the bottleneck of manual annotation, we combine distant supervision from Wikipedia \citelanguageresource{wikipedia} and Wikidata \citelanguageresource{wikidata} with LLM-based quality control, leveraging the ability of current large models to judge annotation consistency even in languages not officially supported.

We hypothesise that certain LLMs can reliably distinguish high-quality annotated sentences from those containing missing or incorrect tags, thus enabling efficient filtering and scalable dataset construction. Our main contributions are as follows:

\begin{enumerate}[(1)]
\item An open pipeline for automatically constructing annotated NER datasets using Wikipedia, Wikidata, and LLM-based judgements.\footnote{\url{https://github.com/plumaj/judgeWEL}}
\item The \texttt{judgeWEL} dataset for Luxembourgish NER, consisting of 28,866 sentences featuring five entity types.\textsuperscript{1}
\end{enumerate}

\section{Related Works}\label{sec:relw}
In this section, we summarise related works in the field of NER. Section \ref{sec:relw1} covers general work, Section \ref{sec:relw2} approaches in low-resource (LR) settings, and Section \ref{sec:relw2-lux} focuses on resources for Luxembourgish. This includes existing language resources, both in terms of data and models.

\subsection{Named Entity Recognition}\label{sec:relw1}
NER is a fundamental Information Extraction and NLP task that identifies and classifies entities (e.g., people, locations, organizations, dates) via sequence labelling schemes such as BIO \cite{ramshaw-marcus-1995-text}. It builds the base for applications like question answering, translation, summarization, and knowledge base construction. 

Research on NER was started with MUC-6 \cite{sundheim1995overview}. Early rule-based systems used linguistic heuristics \cite{mikheev1999named, weischedel1996progress} and gazetteers \cite{rau1991extracting, farmakiotou2000rule, nadeau2006unsupervised}, the latter forming the basis for knowledge-based methods exploiting ontologies \cite{mendes2011dbpedia, rizzo2012nerd}. These approaches required no labelled data and training but lacked (cross-domain) scalability.

Recent work explores generative and LLM approaches. Zero-shot NER with LLMs as standalone remains challenging \cite{xie2023empiricalstudyzeroshotner, hu2024improvinglargelanguagemodels, premasiri2025survey}, but improvements are achieved through prompt engineering and few-shot learning with approaches like PromptNER \cite{ashok2023promptnerpromptingnamedentity}, GL-NER \cite{10.1007/978-3-031-72350-6_30}, and ProML \cite{chen2022promptbasedmetriclearningfewshot}. Hybrid frameworks like SuperICL \cite{xu2023smallmodelsvaluableplugins} and LinkNER \cite{10.1145/3589334.3645414} further integrate LLM reasoning with fine-tuned NER models to enhance accuracy.

\subsection{Named Entity Recognition for Low-Resource languages}\label{sec:relw2}
NER systems for high-resource (HR) languages such as English have achieved state-of-the-art performance due to the availability of large annotated datasets, including CoNLL’03 \cite{tjong2003_conll} and OntoNotes \cite{weischedel2011ontonotes}. However, reaching this performance is difficult in LR settings where labelled data is scarce \cite{ranasinghe-etal-2025-sinhala}. This lack of annotated data is often mitigated through data augmentation and knowledge transfer between HR and LR languages. 

Cross-lingual knowledge transfer aims to transfer learned linguistic and semantic representations from a source language to a target language with limited labelled data. For example, \citet{mayhew-etal-2017-cheap} used translation techniques based on the lexica and Wikipedia to map the knowledge from HR to LR languages. Similarly, \citet{ijcai2018p566} used an extended approach by enriching it with semantic mappings, enabling coverage of named entities unseen in the LR datasets.

\citet{tsai-roth-2016-cross} developed a technique for cross-lingual Wikification, linking non-English Wikipedia entries to English Wikipedia, training multilingual embeddings based on it to facilitate cross-lingual transfer. The WikiAnn dataset \cite{pan-etal-2017-cross} is a large multilingual benchmark covering 282 languages, also created via cross-lingual Wikipedia links. Likewise, Polyglot-NER leveraged distributed word representations encoding semantic and syntactic properties of words across languages, by integrating Wikipedia links and Freebase attributes to create a multilingual NER corpus in 40 languages \cite{doi:10.1137/1.9781611974010.66}.

An intermediate approach between cross-lingual transfer and augmentation is MulDA \cite{liu-etal-2021-mulda}. It addresses the absence of labelled data in a target language by translating limited NER data into multiple target languages using entity placeholders, then generating synthetic multilingual sequences through LMs. 

For low-resource NER augmentation, PromptDA \cite{ijcai2022p590} introduced a label-conditioned word replacement and question-answering style prompting method using BERT. \citet{sym13050786} enhanced LR NER by fine-tuning \texttt{XLM-RoBERTa} using data from HR languages and an attention-based strategy to capture the most relevant semantic and syntactic information. More recent augmentation methods employ LLMs for contextual and entity-level rewriting \cite{ye2024llmdadataaugmentationlarge}, designed specifically for few-shot and low-resource settings. 

Synthetic data generation via prompting LLMs with a small number of gold examples has also shown promising results \cite{kamath2025doessyntheticdatahelp}. The application is further explored in studies examining the effectiveness of LLM-generated synthetic data for NER augmentation.\cite{kamath2025doessyntheticdatahelp}

\subsection{Language Resources for Luxembourgish}\label{sec:relw2-lux}
Only recently has research on NLP for Luxembourgish gained momentum, with a few early exceptions marking the first steps in resource creation and linguistic analysis. \citet{adda-decker-etal-2008-developments} presented initial tools and corpora for computational processing; \citet{snoeren-etal-2010-study} investigated orthographic tendencies such as contextual n-deletion in transcribed speech; and \citet{lavergne-etal-2014-automatic} compiled a manually annotated dataset of mixed-language sentences to test a language identification system. More recently, progress has been made with resources, tasks and models. \citet{sirajzade-etal-2020-sentiment} and \citet{Gierschek2022} developed modern sentiment analysis pipelines, \citet{purschke2020attitudes} proposed an automatic approach for orthographic correction, and \citet{plum2024} started the first treebank for Luxembourgish. \citet{philippy-etal-2024-forget} introduced a zero-shot classification method based on a curated topic dictionary. \citet{ranasinghe-etal-2023-publish} demonstrated a comment moderation task, \citet{lutgen2025} worked on orthographic normalization, and \citet{plum2025} introduced the generative benchmark set \texttt{LuxGen}.

\citetlanguageresource{lothritz-etal-2022-luxembert} contributed a number of classification tasks, including NER. The set was manually annotated for Person, Organisation, Location, Geopolitical Entity, and Miscellaneous tags. The 5500 sentences were extracted from 450 RTL\footnote{The main news broadcaster of Luxembourg.} news articles. Person has the highest number of samples with 2272, Location has the least with 95 samples, and the mean number of occurrences per label is reported at 1214. We refer to this set throughout this paper as \texttt{RTL-NER}.

Various strategies have also been explored for Luxembourgish language models. These include transfer learning from German with \texttt{LuxGPT} \cite{bernardy2022}, data augmentation via generated samples with \texttt{LuxemBERT} \cite{lothritz-etal-2022-luxembert2}, and balanced training data extension via multilingual data for \texttt{LuxT5} \cite{plum2025}. 

Despite these advances, Luxembourgish NLP still remains limited in scope and scale. Resources are still being increased, research activity is gaining momentum, and standard benchmarks remain minimal. Only a few evaluation datasets exist for classification, and some smaller ones cover generative tasks. Expanding these resources, both in terms of data and modelling capacity, is therefore a key objective of the present study.

\section{Dataset Construction}\label{sec:data}
The construction of the \texttt{judgeWEL} (automatically \textbf{judge}d \textbf{W}ikipedia extracted \textbf{E}ntities for \textbf{L}uxembourgish) dataset follows multiple steps. In the following sections, we describe how we select the base data, how we automatically check linked entity types, and how we use LLMs as judges to compile the final dataset. Figure \ref{fig:judgewel-pipeline-onecol} shows an overview of the pipeline.

\begin{figure}[t!]
\centering
\scalebox{0.87}{ 
\begin{tikzpicture}[
    node distance=0.8cm and 0.8cm,
    every node/.style={
        rectangle,
        rounded corners,
        draw=black,
        align=center,
        font=\footnotesize,
        minimum width=6.5cm,
        minimum height=1cm,
        text width=6.3cm
    },
    arrow/.style={->, thick, >=latex}
]

\definecolor{judgegreen}{RGB}{127,200,169} 
\definecolor{finalpurple}{RGB}{163,132,201} 

\node (extract) [fill=blue!7!gray!15] {\textbf{Text Extraction}\\
Luxembourgish Wikipedia\\
\textit{via WikiExtractor}};
\node (link) [below=0.8cm of extract, fill=blue!7!gray!15] {\textbf{Entity Linking}\\
Match hyperlinks to Wikidata\\
BIO tags for PER, ORG, LOC, DATE};
\node (select) [below=0.8cm of link, fill=blue!7!gray!15] {\textbf{Sentence Selection}\\
Skip intro, take next five\\
Filter short or redundant};
\node (annot) [below=0.8cm of select, fill=blue!7!gray!15] {\textbf{Annotation Improvements}\\
\texttt{LuxGPT-NER} + regex\\
Add missing and unify tags};
\node (llmjudge) [below=0.8cm of annot, fill=judgegreen!30] {\textbf{LLM-as-a-Judge}\\
Verify annotation quality\\
Binary keep/discard\\
500-sentence human check (consensus baseline)};
\node (final) [below=0.8cm of llmjudge, fill=finalpurple!25] {\textbf{Final Dataset}\\
\texttt{JudgeWEL}\\
28{,}866 sentences\\
Train/Dev/Test = 80/10/10};

\draw[arrow] (extract.south) -- (link.north);
\draw[arrow] (link.south) -- (select.north);
\draw[arrow] (select.south) -- (annot.north);
\draw[arrow] (annot.south) -- (llmjudge.north);
\draw[arrow] (llmjudge.south) -- (final.north);

\end{tikzpicture}
}

\caption{Overview of the \texttt{JudgeWEL} dataset construction pipeline. Gray stages denote automatic processing, green indicates the LLM-based verification stage, and purple marks the final dataset.}
\label{fig:judgewel-pipeline-onecol}
\end{figure}

\subsection{Base Data}
The data basis is provided by Luxembourgish Wikipedia \citelanguageresource{wikipedia} articles. We work with the 2024-03-29 dump, which features around 77k articles across a variety of topics. While its size is not comparable to the Wikipedias of its neighbouring countries, Luxembourgish articles do follow the other commonalities in terms of how the articles are structured. This structure is the reason why we choose Wikipedia as a source; in particular, we rely on certain entities within the text being hyperlinked to their (existing or non-existing) respective Wikipedia article.

For extracting the articles from the XML dump, we utilise \texttt{WikiExtractor} \citelanguageresource{Wikiextractor2015}. This tool allows us to extract the title and text of each article, while removing any web formatting, such as the info box or index, which are not needed. We extract title and text for each article to json entities, while preserving hyperlink-tagged words in the text, which is important for the next step in the pipeline.

\subsection{Entity Linking} 
Linking entities to their corresponding Wikidata \citelanguageresource{wikidata} page is the next step in the pipeline. Each article is split into sentences using the \texttt{sentence-splitter} \citelanguageresource{sp} package for Python. The splitter is not trained for Luxembourgish, but still performs best when set to German, compared to other options.\footnote{We manually tested a small set of Luxembourgish examples with other options for the language.} For each hyperlinked entity in a sentence, we match the article title via the Wikidata API to check its attributes. More specifically, we check the following attributes for each type of entity that we are targeting:

\begin{description}
    \item[PER (Person)] \textit{P31 = Q5} (human), or has attributes \textit{P569} (birthdate) or \textit{P570} (deathdate).
    \item[ORG (Organisation)] \textit{P31} corresponds to QIDs that denote organisational entities (e.g., \textit{bank}, \textit{organization}, \textit{business}, \textit{hospital}, etc.).
    \item[LOC (Location)] \textit{P31} corresponds to location-type QIDs (e.g., \textit{town}, \textit{municipality}, \textit{country}, etc.).
    \item[DATE (Date)] \textit{P31} equals \textit{Q14795564}, \textit{Q3186692}, or \textit{Q578} (i.e., types of date).
\end{description}

If a hyperlinked entity matches any of the attributes corresponding to a targeted entity type, we annotate the entity tokens with that label. For the labelling, we adopt the BIO (Begin-Inside-Outside) encoding for entity boundary annotation, a format that has been widely used in NER \cite{ramshaw1995_chunking,tjong2003_conll}. Any hyperlinked entity that does not match the targeted entity types, or for which we do not find a corresponding Wikidata entry, is discarded.

\subsection{First Selection Process}
Focusing not only on high-quality but also varied annotated data, we carry out a candidate sentence selection process after all entities have been linked. In order to avoid extremely similar sentence structures, we generally skip the first sentence of each article, as this tends to follow a formulaic pattern introducing the main entity. We then extract the following five sentences, which typically contain the highest concentration of unique hyperlinks. Since Wikipedia conventionally links entities only upon their first mention, these subsequent sentences are more likely to introduce new entities and therefore yield more diverse training examples.

Each sentence is then subjected to a series of quality checks designed to ensure readability, non-overlap between entity spans, and adequate contextual content. Sentences that are too short, entirely capitalised, or composed solely of a single entity mention are excluded. We also remove sentences where entities overlap or occupy nearly the entire span of the sentence, as these tend to produce unreliable labels. To maintain a balanced dataset, we include at most one negative example (a sentence containing no entities) per article, provided that it meets minimal length criteria. Duplicate sentences are also filtered out.

This selection procedure results in a pool of candidate sentences that are both linguistically varied and structurally sound, providing a robust basis for the subsequent LLM-based verification stage. In doing so, we prioritise diversity and contextual completeness over maximal volume, ensuring that later filtering and evaluation steps operate on genuinely informative examples rather than redundant or low-quality material.

\begin{figure}[h!]
\centering
\fbox{%
\begin{minipage}{0.9\linewidth}
\scriptsize
\setlength{\parskip}{2pt}

\texttt{\\
\textcolor{blue!60!black}{\textquotedbl id\textquotedbl}:\textcolor{brown!70!black}{\textquotedbl 1/100-1\textquotedbl},\\
\textcolor{blue!60!black}{\textquotedbl text\textquotedbl}:\textcolor{brown!70!black}{\textquotedbl De Jhempi Kniddel schafft beim Staat\textquotedbl},\\
\textcolor{blue!60!black}{\textquotedbl tokens\textquotedbl}:[%
\textcolor{brown!70!black}{\textquotedbl De\textquotedbl}, 
\textcolor{brown!70!black}{\textquotedbl Jhempi\textquotedbl}, 
\textcolor{brown!70!black}{\textquotedbl Kniddel\textquotedbl}, 
\textcolor{brown!70!black}{\textquotedbl schafft\textquotedbl}, 
\textcolor{brown!70!black}{\textquotedbl beim\textquotedbl}, 
\textcolor{brown!70!black}{\textquotedbl Staat\textquotedbl}, 
\textcolor{brown!70!black}{\textquotedbl .\textquotedbl}],\\
\textcolor{blue!60!black}{\textquotedbl labels\textquotedbl}:[%
\textcolor{brown!70!black}{\textquotedbl O\textquotedbl}, 
\textcolor{brown!70!black}{\textquotedbl B-PER\textquotedbl}, 
\textcolor{brown!70!black}{\textquotedbl I-PER\textquotedbl}, 
\textcolor{brown!70!black}{\textquotedbl O\textquotedbl}, 
\textcolor{brown!70!black}{\textquotedbl O\textquotedbl}, 
\textcolor{brown!70!black}{\textquotedbl B-ORG\textquotedbl}, 
\textcolor{brown!70!black}{\textquotedbl O\textquotedbl}]\\
}

\end{minipage}}
\caption{JSON representation of a labelled sentence used for the dataset.}
\label{fig:annotation-json-example}
\end{figure}

For the project at hand, the selection process results in 74,710 annotated sentences after the initial selection. Certain samples show unlabelled, mislabelled, and incomplete entities, as well as sentences that should not contain any entities (as negative examples) but do contain entities. After this selection, all sentences are stored in JSON format, as depicted by the example in Figure \ref{fig:annotation-json-example}. Each sentence is given a unique identifier, consisting of a running number, the Wikipedia article ID that the sentence originates from, and the sentence number within that article.

\subsection{Annotation Improvements}
Since the Luxembourgish Wikipedia does not cover all entities present in our dataset due to its size, we perform an additional automatic annotation step. Using a \texttt{LuxGPT} model fine-tuned  through prompt engineering for NER with a modified version of \texttt{RTL-NER} dataset \cite{bernardy2022}, we re-evaluate all entities initially labelled with O-tags and, where necessary, reassign them with one of the predefined entity tags. Some manual post-processing is conducted to harmonize the tag set, including the unification of the GPE tag under LOC. This is because the \texttt{RTL-NER} distinguishes between these two tags (see Section \ref{sec:relw2-lux}), whereas we do not. These adjustments are implemented using simple regular expressions.

As both annotation procedures occasionally result in insufficient tagging of date expressions, we subsequently apply a regular expression–based script that searches for date patterns within token sequences. These patterns include both numerical formats and Luxembourgish expressions used to indicate temporal information.

\subsection{LLM-as-a-Judge}
To minimise unlabelled, mislabelled, or incomplete entities in the dataset, we test several LLMs to verify each candidate sentence. This stage involves iterative prompt testing to identify those that best retain correctly annotated examples while filtering out inconsistent ones. Given the already substantial size of the automatically derived corpus, our aim is not exhaustive coverage but reliability: we deliberately accept the loss of some valid samples in favour of higher overall consistency. The final prompt used for verification is shown in Figure \ref{fig:annotation-prompt}. The resulting instruction is concise but explicit, guiding the model towards binary and decisive judgements.

\begin{figure}[h!]
\centering
\fbox{%
\begin{minipage}{0.9\linewidth}
\footnotesize 
\setlength{\parskip}{4pt} 
\textbf{Prompt:}
For each sentence below, please check whether the named entity labels are appropriate. 
Each sentence will be presented as a list of tokens, and you will be provided with a list of BIO tags that indicate the entities. 
Some tokens may be labelled with just a tag and no BIO if they are a singular item; that is acceptable. 
Some sentences will have no labelled items at all and may be examples of no labels applying, a negative example. 
The accepted labels are \texttt{PER} for a person, \texttt{ORG} for an organisation, \texttt{LOC} for a location, \texttt{DATE} for a specific date or year, and \texttt{MISC} for any other named entity that does not fit any of the other types.

Your task is to check each sentence: make sure the suggested labels are correct, make sure that all potential labels have been assigned, and that labels have been assigned the correct BIO range. 
For unlabelled sentences, you need to make sure that these are valid negative examples, so that really no labels apply. 
Since you will be filtering the sentences, you will have to be decisive, offering a \texttt{1} (yes, keep sentence) or \texttt{0} (no, discard sentence) answer.

\end{minipage}}
\caption{LLM-as-a-Judge prompt used.}
\label{fig:annotation-prompt}
\end{figure}

We also include a system message for each prompt that takes care of formatting, opting for a CSV output containing the sentence ID and the verification label produced by the model. Sentences are passed in batches of 20 in the previously described JSON format. 

Once the prompt is refined to a satisfactory level, we evaluate a diverse set of LLMs to identify the most reliable judge for Luxembourgish data. The model selection balances open and closed systems, scale regimes, and ecosystem diversity, thereby strengthening empirical validity and reproducibility:

\paragraph{Proprietary Baselines.}
\texttt{GPT-5} represents the cutting edge of commercial models and serves as a reference benchmark for high-end proprietary performance \citep{openai2025_introducing_gpt5}. Its smaller counterpart, \texttt{GPT-5-mini}, provides a more lightweight and cost-efficient variant for comparison, enabling us to assess performance scalability within the same model family.

\paragraph{Open-Weight Systems.}
\texttt{Gemma-3-27B-IT} is included because Google/DeepMind released it under a permissive licence, allowing transparent evaluation and reproducibility. \texttt{Mistral-Medium-3.1} represents the new generation of mid-sized open models optimised for efficiency and multilingual generalisation, offering a valuable test case for open community-led development. Finally, \texttt{GPT-OSS-120B} acts as a large open-weight counterpart to the closed models, released by OpenAI under Apache 2.0 with near-parity on reasoning benchmarks, providing a direct comparison point for open vs. proprietary large-scale systems \citep{openai2025_gptoss_model_card, openai2025_introducing_gpt_oss}.

\paragraph{Instruction-Tuned Systems.}
\texttt{LLaMA-3.3-8B-Instruct} offers a smaller, instruction-optimised model within a widely adopted open-weight ecosystem \citep{touvron2024_llama3}. In addition, \texttt{command-a-03-2025} represents a new generation of instruction-tuned commercial models that occupy the middle ground between fully open and closed systems, and is said to emphasise controlled outputs and multilingual understanding \citep{cohere2025_command_a}.

Taken together, these models cover the current LLM landscape: closed vs. open, large vs. compact, and general-purpose vs. instruction-tuned. Covering these areas allows for a balanced and comprehensive evaluation of model behaviour across technological and governance regimes.

\begin{table*}[ht!]
\centering
\small
\begin{tabular}{l|r|rrrrr}
\toprule
\textbf{Split} & \textbf{Sents.} & \textbf{PER} & \textbf{ORG} & \textbf{LOC} & \textbf{DATE} & \textbf{MISC} \\
\midrule
\texttt{Train} & 22{,}947 & 11{,}013 & 2{,}620 & 10{,}341 & 11{,}355 & 243 \\
\texttt{Dev}   & 2{,}868  & 1{,}490 & 382   & 1{,}367 & 1{,}414 & 61 \\
\texttt{Test}  & 3{,}051  & 1{,}251 & 304   & 1{,}310 & 1{,}523 & 13 \\
\bottomrule
\end{tabular}
\caption{Total sentences and entity type counts for the \texttt{judgeWEL} data splits.}
\label{tab:dataset-stats}
\end{table*}

\subsubsection{Human-as-a-Judge}
To verify the LLM-judged sentences, and also keep human-oversight to a minimum, two annotators (Luxembourgish native speakers, aged between 20-30) were tasked to verify 500 sentences of the dataset. The annotators judged each sentence according to the prompt for the LLMs, so that maximum comparability was retained. The annotators agreed 414 out of 500 times, with Cohen's $\kappa$ at 0.66. The annotation process took 6 hours per annotator. For each disagreement, a consensus was reached, which we consider the final human label. Of the 500 sentences, 182 were verified as high-quality examples, and added to the test set.

After human annotation, we evaluate the LLMs on the same set to select the best. The evaluation of the LLMs as a judge is presented in Section \ref{sec:eval-llms}. 

\subsection{Final Dataset}
We compile \texttt{judgeWEL} after obtaining the LLM judgements. In total, this brings down the number of sentences to 28,866. For the training, development, and testing splits we opted for roughly 80\%, 10\% and 10\%. We shuffle the data, and obtain the label balances as shown in Table \ref{tab:dataset-stats}. Overall, PER is the most common tag, followed by DATE and LOC. For ORG, we obtain far fewer samples, and very few for MISC.

\section{Evaluation}\label{sec:models}
This section evaluates both the construction and use of the \texttt{judgeWEL} dataset as described in the previous section. Specifically, we evaluate the ability of LLMs to act as reliable judges of annotation quality (Section \ref{sec:eval-llms}), and the downstream performance of models supporting Luxembourgish on the NER task (Section \ref{sec:eval-ner}).

\subsection{LLM-as-a-Judge}\label{sec:eval-llms}
Table \ref{tab:model-agreement} presents the agreement between each LLM and the human annotator consensus. For each model, we report the absolute number of matching labels and Cohen’s $\kappa$ as a measure of inter-rater reliability, alongside the inter-annotator agreement between humans for reference.

\begin{table}[h!]
\centering
\small
\begin{tabular}{lcc}
\toprule
\textbf{Model} & \textbf{Agreement} & \textbf{$\kappa$} \\
\midrule
\texttt{Gemma-3-27B-IT} & 177 & -0.29 \\
\texttt{LLaMA-3.3-8B-Instruct} & 236 & -0.05 \\
\texttt{command-a-03-2025} & 349 & 0.40 \\
\texttt{Mistral-Medium-3.1} & 362 & 0.45 \\
\texttt{GPT-OSS-120B} & 367 & 0.47 \\
\texttt{GPT-5(-mini)} & 405 & 0.62 \\
\midrule
\textit{Human Annotators} & \textit{414} & \textit{0.66} \\
\bottomrule
\end{tabular}
\caption{Model agreement with annotator consensus labels on 500 test samples.}
\label{tab:model-agreement}
\end{table}

Overall, \texttt{GPT-5-mini} and \texttt{GPT-5} produce the exact same output, and clearly outperform all other systems, reaching a $\kappa$ value of 0.62, approaching the level of human inter-annotator agreement at 0.66. This indicates that high-end proprietary models can understand Luxembourgish text and capture the fine-grained cues that distinguish well-annotated from inconsistent examples. The fact that the smaller model produces the same output as the larger model also has significant cost implications, as the former cost roughly \$25 to annotate the entire pre-filtered set of 74k sentences, while the latter would have cost around \$180. Among the open-weight systems, \texttt{GPT-OSS-120B} and \texttt{Mistral-Medium-3.1} perform comparatively well, showing moderate and stable alignment with human judgements. The \texttt{command-a-03-2025} model achieves slightly lower agreement but remains consistent across samples, suggesting solid comprehension but a less decisive evaluative behaviour. By contrast, \texttt{Gemma-3-27B-IT} and \texttt{LLaMA-3.3-8B-Instruct} exhibit near-zero or negative correlation with human ratings, confirming that instruction tuning alone does not ensure evaluative reliability in smaller open-weight systems.

A finer-grained analysis of errors broken down by entity type reveals that these aggregate differences mask systematic per-category biases, evaluated on the 388 sentences for which both human annotators agreed. DATE-containing sentences are the most consistently handled across all models, with every system achieving its highest or near-highest F1 on this category, likely because temporal expressions provide reliable surface cues that LLM judges can exploit regardless of their general evaluative quality. PER entities show a wider spread: \texttt{Mistral-Medium-3.1} and \texttt{command-a-03-2025} both exceed F1 72, while \texttt{GPT-OSS-120B} drops to 39.2, driven by a strong conservative bias that causes it to discard 52 of the 71 gold-keep PER sentences, a pattern consistent with its overall tendency to over-reject. MISC entities prove the most problematic across the board: \texttt{GPT-5} and \texttt{GPT-OSS-120B} achieve an F1 of zero, never retaining a MISC sentence that human annotators approved, while the remaining models reach at best 17, largely driven by false positives. This likely reflects the heterogeneous and language-specific nature of the MISC class in Luxembourgish NER. Finally, sentences containing no named entities expose the sharpest divide: \texttt{GPT-5} and \texttt{GPT-OSS-120B} handle these near-perfectly (F1 99.2 and 97.4), while \texttt{Gemma-3-27B-IT} discards all 60 such sentences, yielding an F1 of zero. This reinforces the interpretation that its negative $\kappa$ reflects a systematic tendency to equate the absence of named entities with annotation failure, rather than random disagreement.

Taken together, these results show that while proprietary models continue to lead in judgement accuracy, large open-weight and mid-size commercial models are closing the gap, which indicates promising potential for transparent, reproducible evaluation pipelines in the near future.

\subsection{Named Entity Recognition Task}\label{sec:eval-ner}
For the NER task, we benchmark a set of multilingual and Luxembourgish-specific encoder models: \texttt{mBERT} \cite{devlin2018}, \texttt{LuxemBERT} \citelanguageresource{lothritz-etal-2022-luxembert}, and \texttt{XLM-RoBERTa} \cite{conneau2019}. Alongside this, we fine-tune and benchmark three generative LLMs: \texttt{LuxGPT-NER},  \texttt{aya-expanse-8b} \cite{2024ayamodelinstructionfinetuned} and \texttt{Meta-Llama-3-8B-Instruct} \citep{touvron2024_llama3}. We also tested \texttt{GPT-5} in a zero- and few-shot setting, however, the model did not deliver consistent results (producing just O tags) and we do not discuss any further results, as finding the optimal model for this task was not our objective.

The classification models are trained with the same settings in all training/testing combinations, as the aim was to comparative benchmarking above maximum performance. We used the following hyperparameters: 3 epochs, input length of 128, and a learning rate of $5e-5$. All classification models were trained on the same server. The fine-tuned LLMs were trained for 5 epochs with Early Stopping, which occured for \texttt{LuxGPT-NER} after the first, for \texttt{Llama} and \texttt{Aya} after the third epoch. The other hyperparameters used were a learning rate of $2e-5$ and an input length of maximum 512. Due to the LLM architecture, the NER-tagging task was altered into a sequence modelling task, with the token sequence to tag as prompt to complete with the sequence of respective tags. \texttt{Aya} and \texttt{Llama} were also given an additional system prompt defining the task. For the evaluation, the token sequences were compared to the test sets.

Evaluation scores are obtained for all models using the \texttt{seqeval} \citelanguageresource{seqeval} package for Python, and we report the micro averaged scores rounded to two decimals for each metric per model

To understand the potential performance increase achievable with \texttt{judgeWEL}, we carry out the evaluation in two settings. First, we train, validate and test on \texttt{judgeWEL} (Section \ref{sec:eval-new}), constructed with the methodology laid out in this paper. Second, we train and validate on \texttt{judgeWEL}, but test using the existing \texttt{RTL-NER} dataset for Luxembourgish (Section \ref{sec:eval-old}), which we previously described in Section \ref{sec:relw2-lux}.

\subsubsection{judgeWEL}\label{sec:eval-new}
As shown in Table \ref{tab:model-eval}, all transformer-based encoders perform well, reaching F1-scores above 0.90. \texttt{LuxemBERT} slightly outperforms multilingual baselines, confirming that language-specific pretraining yields measurable gains even on relatively small corpora. \texttt{XLM-RoBERTa} follows closely, demonstrating strong cross-lingual transfer from related languages such as German. These results establish a reliable baseline for Luxembourgish NER.

\begin{table}[h!]
\centering
\small
\begin{tabular}{lccc}
\toprule
\textbf{Model} & \textbf{Precision} & \textbf{Recall} & \textbf{F1-score} \\
\midrule
\texttt{LuxGPT-NER} & 0.63 & 0.74 & 0.68 \\
\texttt{Aya-exp-8b} & 0.86 & 0.83 & 0.84 \\
\texttt{LLaMA-3-8B} & 0.93 & 0.91 & 0.92 \\
\texttt{mBERT}      & 0.92 & 0.90 & 0.91 \\
\texttt{XLM-R}      & 0.92 & 0.91 & 0.91 \\
\texttt{LuxemBERT}  & 0.92 & 0.92 & 0.92 \\
\bottomrule
\end{tabular}
\caption{Evaluation results for each model trained and tested on the \texttt{judgeWEL} dataset.}
\label{tab:model-eval}
\end{table}

Among the generative LLMs, performance varies more widely. \texttt{Meta-Llama-3-8B-Instruct} achieves the same performance level as LuxemBert with a F1 score of 0.92. \texttt{Aya-expanse-8b} achieves an F1-score of 0.84, which is slightly below the encoder-based baselines, but still indicates strong generalisation and semantic awareness of entity boundaries. In contrast, the fine-tuned \texttt{LuxGPT-NER} reaches 0.68, suggesting that despite adaptation to Luxembourgish, it struggles to maintain consistent token-level alignment during generation. Manual inspection confirms that both models occasionally merge tokens, omit boundaries, or produce malformed BIO sequences. This points to a broader limitation of autoregressive architectures for sequence-labelling tasks, where structured, token-aligned predictions are required. The same tendency was observed in our zero-shot \texttt{GPT-5} evaluation, where high-level understanding did not translate into stable, fine-grained labelling.

\begin{figure*}[htp!]
\centering
\includegraphics[scale=0.7]{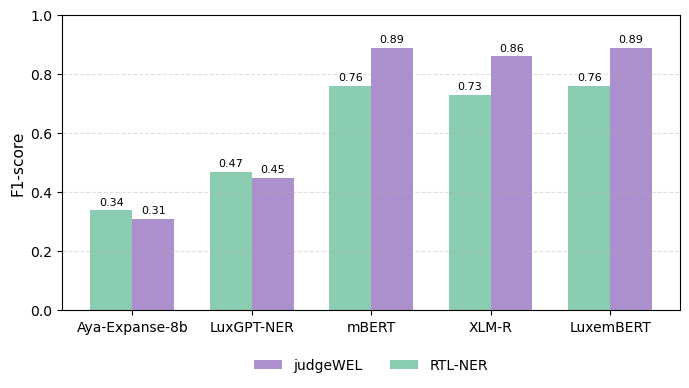}
\caption{Model F1 scores trained on \texttt{judgeWEL} or \texttt{RTL-NER}, and evaluated on \texttt{RTL-NER}.}
\label{fig:eval-comp}
\end{figure*}

\subsubsection{Comparison with RTL-NER}\label{sec:eval-old}
To compare performance gains that our dataset brings to Luxembourgish NER, we evaluate the same set of models in two further comparative scenarios. First, we train on the \texttt{judgeWEL} dataset and evaluate on \texttt{RTL-NER}. To ensure comparability, \texttt{judgeWEL} is downsampled to match the size of the \texttt{RTL-NER} training split using stratified sampling. We then train and evaluate on \texttt{RTL-NER} directly. Because the existing set features different labels, and to ensure maximum comparability, we combine the GPE and LOC entity types into one. The existing set does not feature the DATE entity type, for which we make no changes. Second, we retrain the models using the entire existing dataset, but again with GPE and LOC entity types combined.

\begin{table}[h!]
\centering
\small
\begin{tabular}{lccc}
\toprule
\textbf{Model} & \textbf{Precision} & \textbf{Recall} & \textbf{F1-score} \\
\midrule
\texttt{LuxGPT-NER} & 0.36 & 0.62 & 0.45 \\
\texttt{Aya-exp-8b} & 0.29 & 0.34 & 0.31 \\
\texttt{LLaMA-3-8B} & 0.02 & 0.01 & 0.02 \\
\texttt{mBERT}      & 0.93 & 0.91 & 0.92 \\
\texttt{XLM-R}      & 0.92 & 0.91 & 0.91 \\
\texttt{LuxemBERT}  & 0.92 & 0.92 & 0.92 \\
\bottomrule
\end{tabular}
\caption{Evaluation results for each model trained on \texttt{judgeWEL} and evaluated on \texttt{RTL-NER}.}
\label{tab:model-eval-2}
\end{table}

When classification models are trained on the new dataset and evaluated on the existing one, performance remains consistently high across architectures, with \texttt{mBERT} and \texttt{LuxemBERT} achieving F1-scores of 0.89, and with \texttt{XLM-RoBERTa} at 0.86 (see Table \ref{tab:model-eval-2}). This indicates that the automatically constructed and judged dataset captures entity boundaries accurately and generalises well to human-annotated data in a different domain. The stability of cross-dataset performance suggests strong label alignment and coverage across entity types, despite the partial mismatch in label inventories. Model performance increases to around 0.91 in terms of F1 score when trained on the full training set, which is roughly 4 times larger.

By contrast, classification models trained solely on the original \texttt{RTL-NER} dataset reach substantially lower F1-scores, confirming that the new corpus provides broader contextual diversity and higher-quality supervision. Figure \ref{fig:eval-comp} shows the differences in F1 score per model.

When comparing the results of the generative LLMs across setups, the results change. We observe a small but consistent drop in performance when moving from in-domain training on \texttt{RTL-NER} (0.34 for \texttt{Aya-expanse-8b}, 0.47 for \texttt{LuxGPT-NER}) to cross-domain evaluation using \texttt{judgeWEL} for training (0.31 and 0.45, respectively). This decline suggests limited transferability of token-level labelling behaviour across datasets, even though entity semantics are largely preserved. Closer inspection reveals that both models tend to generate incomplete or inconsistent BIO sequences, frequently omitting boundaries or merging adjacent entities. For \texttt{Meta-Llama-3-8B-Instruct} this effect shows especially strong in a complete performance drop. The issue appears structural rather than linguistic: autoregressive LLMs are not optimised for deterministic, token-aligned output, and thus struggle to reproduce the rigid labelling constraints that encoder-based architectures handle naturally. The same pattern was already visible in the zero-shot \texttt{GPT-5} evaluation, where the model exhibited unstable tagging. 

Overall, the results highlight the robustness of our approach: models trained on automatically generated data can transfer effectively to existing human-labelled benchmarks, thereby reducing the dependency on costly manual annotation. This validates the use of LLM-assisted distant supervision for future corpus creation in comparable low-resource languages. The findings also underline that while instruction-tuned LLMs capture entity meaning effectively, transformer encoders remain more robust for precise sequence-labelling, particularly when generalising beyond their training domain.

\section{Conclusion}\label{sec:conc}
This study set out to examine whether LLMs can reliably assist in the construction and validation of an NER dataset for Luxembourgish, a language with limited but growing annotated resources. The findings confirm that current LLMs can serve as effective judges of annotation quality, reaching levels of agreement close to human inter-annotator consensus. This supports our hypothesis that LLMs are able to distinguish between high and low-quality examples, even when the language in question is not explicitly supported during pre-training.

At the same time, our results suggest that fully automating label generation through LLMs remains premature. While models such as \texttt{GPT-5} display strong discriminative capabilities, they still introduce systematic inconsistencies when tasked with producing entity labels from scratch. The hybrid approach proposed here, which combines structured resources such as Wikipedia and Wikidata with selective LLM-based filtering, thus represents a pragmatic middle ground between manual and fully generative annotation.

Future work will focus on extending this methodology to other under-represented languages with similar data profiles, as well as on refining label granularity to capture additional entity subtypes and relational information. Incorporating limited human-in-the-loop validation and cross-lingual alignment could further enhance both precision and transferability. Ultimately, this work demonstrates that careful use of LLMs, rather than full automation, offers a sustainable path towards equitable NLP. And for languages with little data, a combination of structured knowledge and model verification may prove more powerful than either alone.

\section{Acknowledgements}
The data annotation for evaluation purposes was carried out by student assistants, funded by the Horizon Europe project LLMs4EU. We thank the annotators for their work.

The computational experiments reported in this paper were conducted on the MeluXina high-performance computing infrastructure, an allocation granted by the University of Luxembourg on the EuroHPC supercomputer hosted by LuxProvide.

Tharindu Ranasinghe is partially funded from COST Action CA23147 GOBLIN – Global Network on Large-Scale, Cross-domain and Multilingual Open Knowledge Graphs, supported by COST (European Cooperation in Science and Technology).

\section{Ethical Considerations}
The dataset introduced in this paper is derived from publicly available sources, primarily Wikipedia and Wikidata, which are both governed by permissive licences (CC BY-SA 4.0 and CC0 respectively). No private or sensitive user information was accessed or processed, and all data handling complied with the applicable terms of use. The automatic annotation process may nonetheless propagate biases present in the original sources, such as uneven entity representation across gender, geography, or topic. While these issues are partially mitigated by the scale and diversity of Wikipedia, they warrant explicit acknowledgement and monitoring in downstream applications.

The use of LLMs as annotation judges also introduces potential ethical concerns. Proprietary models such as \texttt{GPT-5} are not fully transparent in their training data or alignment processes, limiting interpretability and reproducibility. To balance this, we included a set of open-weight models to ensure that the pipeline remains inspectable and reproducible. The dataset and code will be released under open licences to support transparency, reproducibility, and responsible research on low-resource languages.

All human annotators involved in this study were compensated for their work in accordance with national wage standards and institutional ethics policy.

\section{Limitations}
While the proposed approach substantially reduces the cost of data creation for under-represented and under-resourced languages, several limitations remain. First, the methodology depends on the existence of a sufficiently populated Wikipedia edition and reliable entity linking to Wikidata. For languages with extremely sparse or inconsistent Wikipedia coverage, this strategy may yield limited or biased samples. Second, although LLMs can approximate human judgement, their evaluations are still probabilistic and subject to variation across prompts, temperature settings, and API versions. 

Furthermore, the dataset focuses on five coarse-grained entity categories, which may not capture finer semantic distinctions (e.g., between organisations and institutions or between temporal and numerical expressions). Finally, while the resulting dataset improves lexical and topical coverage for Luxembourgish, its automatic origin may still include residual annotation noise. Manual validation or iterative human-in-the-loop refinement will be essential for tasks requiring high precision, such as information extraction in sensitive or policy-related domains.

\section{Bibliographical References}\label{sec:reference}
\bibliographystyle{lrec2026-natbib}
\bibliography{refs}

\section{Language Resource References}
\label{lr:ref}
\bibliographystylelanguageresource{lrec2026-natbib}
\bibliographylanguageresource{resources}

\end{document}